\documentclass{article}
\usepackage{spconf}
\usepackage{amsthm}
\usepackage{amsmath, amssymb, amsfonts}
\usepackage{graphbox}
\usepackage{multirow}
\usepackage{hyperref}
\hypersetup{hidelinks}
\usepackage{cite}
\usepackage{algorithmic}
\usepackage{graphicx}
\usepackage{textcomp}
\usepackage{xcolor}

\begin{document}

\name{Valentin Penaud-{}-Polge, Santiago Velasco-Forero, Jesus Angulo\thanks{This work was granted access to the HPC resources of IDRIS under the allocation 2022-[AD011013367] made by GENCI. This work has been supported by Fondation Math\'ematique Jacques Hadamard(FMJH) under the PGMO-IRSDI 2019 program.}}

\address{Mines Paris, PSL University\\
		Center for Mathematical Morphology\\
		\{valentin.penaud\_polge, santiago.velasco, jesus.angulo\} @minesparis.psl.eu}

\title{Fully Trainable Gaussian Derivative Convolutional Layer}


\maketitle

\begin{abstract}
	The Gaussian kernel and its derivatives have already been employed for Convolutional Neural Networks in several previous works. Most of these papers proposed to compute filters by linearly combining one or several bases of fixed or slightly trainable Gaussian kernels with or without their derivatives.
	In this article, we propose a high-level configurable layer based on anisotropic, oriented and shifted Gaussian derivative kernels which generalize notions encountered in previous related works while keeping their main advantage.
	The results show that the proposed layer has competitive performance compared to previous works and that it can be successfully included in common deep architectures such as VGG16 for image classification and U-net for image segmentation.
\end{abstract}


\section{Introduction}
\label{sec:RW}
The Gaussian kernel takes a major place in the image processing literature due to its proximity with the visual system and its scale-space representations. Several works have been published to determine the most adapted family of filters for scale-space representation by assuming constraints. All converged towards the Gaussian kernel and its derivatives to form the local $N$-jet \cite{florack1992scale, florack1996gaussian, lindeberg2011generalized, koenderink1984structure, koenderink1987representation}. Moreover, the Gaussian kernel coincides with the receptive field profiles of the human front-end visual system and from a mathematical viewpoint, the local $N$-jet is fundamental as it represents a local Taylor approximation of the studied image. This makes the local $N$-jet an efficient tool for image representation and description \cite{larsen2012jet, manzanera2011local}.

For Convolutional Neural Networks, several approaches have been published considering either only the Gaussian kernel \cite{tabernik2018spatially, shelhamer2019blurring} or the Gaussian kernel together with its derivatives \cite{jacobsen2016structured, lindeberg2020scalegaussnet, sundaramoorthi2019translation}. The layers proposed in these references have the advantage to make the size of the filters independent of the number of parameters. Tabernik et al. \cite{tabernik2018spatially} proposes a convolutional layer whose filters are defined as mixtures of isotropic Gaussian kernels, also called mixture of displaced aggregation units (DAU). The proposed layer learns the mean of each Gaussian kernel and its weight in the mixture. The number of DAU in each filter is left as a hyperparameter and the scale of the DAU is either fixed \cite{tabernik2018spatially} or learned by the network \cite{tabernik2016towards, tabernik2020spatially}. In \cite{shelhamer2019blurring}, it is proposed to compose free-form filters with Gaussian kernels to benefit from the diversity of variations of the free-form filters while adapting them to the shape of the structured filters. Nevertheless, as free-form filters are used, the number of parameters is similar to conventional convolutional layers.
\begin{figure*}[t]
	\centering
	\bgroup
    \def\arraystretch{0.6}
	\begin{tabular}{cc}
		\includegraphics[align=m,width=0.26\linewidth]{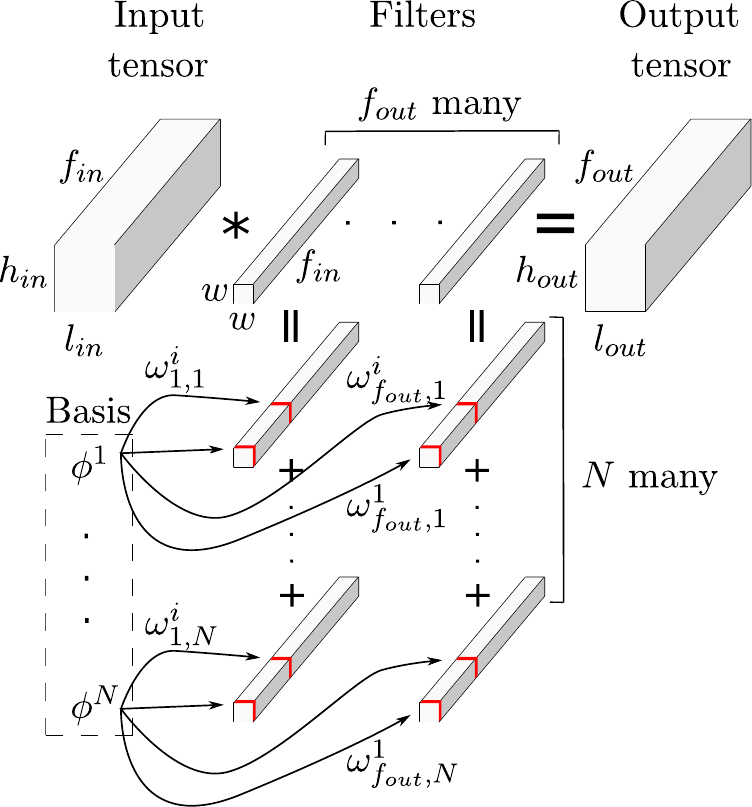}& \hspace{1cm}
		\includegraphics[align=m,width=0.49\linewidth]{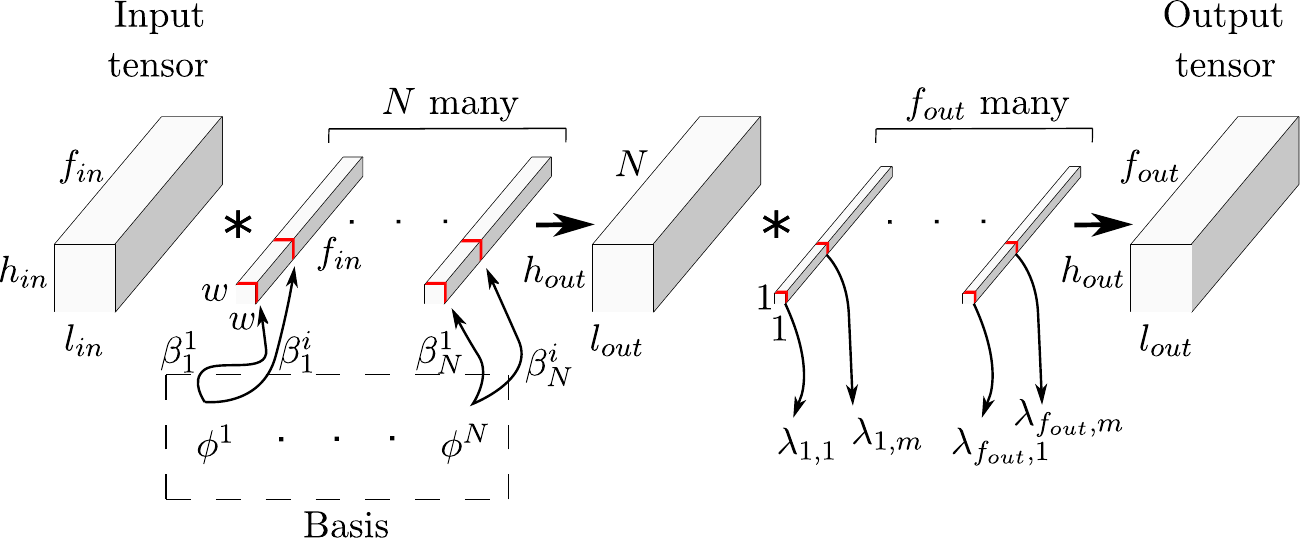} \\
		(a) & \hspace{1cm} (b)
	\end{tabular}
	\egroup
	\caption{(a): Schematic illustration of eq. \eqref{eq6}. (b): Schematic illustration of eq. \eqref{eq9}.}
	\label{fig1}
\end{figure*}
In \cite{jacobsen2016structured}, a layer using kernels defined to be linear combinations of a basis of isotropic Gaussian derivative kernels is proposed. More precisely, the basis of each layer is composed of local $N$-jets. Only the weights of the linear combinations are learned by the network while the order of the $N$-jets, their scales and their orientations are fixed and considered as hyperparameters. The proposed type of neural network is called Structured Receptive Field Neural Network (RFNN). The advantage of RFNN is shown when a small amount of training data is available as the number of parameters is reduced in comparison with standard CNN. RFNN has been reused in \cite{sundaramoorthi2019translation} to focus on the translation insensitivity of this network where the Gaussian derivative kernels incorporate Lipschitz continuity. Recently, a version of RFNN where the network learns the scales of the isotropic, non-oriented and non-shifted Gaussian derivative kernels has been published \cite{pintea2021resolution}. Finally, Lindeberg \cite{lindeberg2020scalegaussnet} presents a scale invariant neural network whose filters are similar to those used in \cite{jacobsen2016structured}. To obtain scale in(equi)variance, Lindeberg reuses, in a modified manner, the principle of scale-channel networks \cite{jansson2020exploring}.

None of these references investigated the possibility of a network learning all the parameters encountered in a mixture of anisotropic, shifted and oriented Gaussian derivatives. Moreover in all these references, some proposed layers have filters sharing a unique basis of Gaussian derivative kernels \cite{jacobsen2016structured} for several coefficients of the linear combination, some proposed layers have filters sharing coefficients of the linear combinations with different scaled bases \cite{lindeberg2020scalegaussnet} and some layers uses one basis for each filter \cite{tabernik2018spatially}.

\newpage The aim of this paper is also to propose a set of layers whose filters can share a basis of Gaussian derivatives or have their own individual basis or "in-between" configurations. We will also propose two ways to linearly combine the Gaussian derivative kernels when dealing with multi-channels inputs.
\section{Trainable Gaussian Derivative Layers}
\label{sec:FLGDL}
In this section we present the fully trainable Gaussian derivative convolutional layers. The layers are defined in a way that it is based on the local $N$-jet like RFNN and Lindeberg's Gaussian derivative network to restrain the number of parameters. They also offer more complex filters, like it is the case in \cite{shelhamer2019blurring} but without the use of free-form filters. Finally they have adapted receptive fields as is the case in \cite{tabernik2018spatially} and they generalize the notion of number of basis. First, we will recall the link between Gaussian derivatives and Hermite polynomials as it constitutes a convenient way of parameterising a deep learning convolution kernel into an anisotropic, shifted and oriented Gaussian derivative kernel.

\bigskip
\noindent \textbf{Hermite Polynomials and Gaussian Derivatives.}
The one-dimensional centered Gaussian kernel and its $p^{th}$ derivative are respectively given by 
$G\left(x, \sigma \right) =$ $\frac{1}{\left(2\pi\sigma^2\right)^{\frac{1}{2}}} e^{\frac{-x^2}{2\sigma^2}}$  and $G_p\left(x, \sigma \right) =$ $\frac{\partial^{p}}{\partial x^p}G\left(x, \sigma\right)$,
where $\sigma$ represents the standard deviation (or scale). 
Gaussian derivatives of any order can be expressed as a Hermite polynomial multiplied by a Gaussian kernel \cite{yang2013steerability}, and as Hermite polynomials are easily expressible, we will use them to formulate the anisotropic Gaussian derivative kernels without any use of derivation operations. The one-dimensional Hermite polynomial of order $p$ is given by the following equation expressed in a form of a serie which makes it easier to use in practice
$H_p\left(x\right) =$ $\sum_{i = 0}^{\lfloor p/2 \rfloor} \frac{\left(-1\right)^i p!}{i! \left(p - 2i\right)!}\left(2x\right)^{p-2i}$.
The relation between the one-dimensional Gaussian derivative of order $p$ and the Hermite polynomial of order $p$ is given by
$G_p\left(x, \sigma\right) =$ $\left(-\frac{1}{\sqrt{2}\sigma}\right)^p\sqrt{2 \pi}\sigma H_p\left(\frac{x}{\sqrt{2}\sigma}\right)G\left(\frac{x}{\sqrt{2}}, \sigma\right)^2$.
From this, we can define the two-dimensional anisotropic Gaussian derivative kernel as follow:
$G_{p,q}\left(x_1, x_2, \sigma_{x_1}, \sigma_{x_2}\right) =$ $ G_p\left(x_1, \sigma_{x_1}\right) G_q\left(x_2, \sigma_{x_2}\right).$
For convenience, we will use the notation $\mathbf{x} = \left(x_1, x_2\right)$ and $\boldsymbol{\sigma} = \left(\sigma_{x_1}, \sigma_{x_2}\right)$. The resulting filters can be oriented by an angle $\theta$ and shifted by $\boldsymbol{\mu}$ giving
\begin{equation}
\label{eq4}
G_{p, q} \left(\mathbf{u}, \boldsymbol{\mu}, \boldsymbol{\sigma}, \theta \right) =  G_p\left(y_1, \sigma_{u_1}\right)G_q\left(y_2, \sigma_{u_2}\right),
\end{equation}
with $\left(y_1, y_2\right) = \left(u_1 - \mu_1, u_2- \mu_2\right)$ and 
$$\left(u_1,u_2\right) = \left(x_1 \cos\theta + x_2 \sin\theta, -x_1 \sin\theta + x_2 \cos\theta \right)$$. The bases used in the proposed layer are defined and denoted as $\lbrace \phi^1, ..., \phi^N \rbrace = \lbrace G_{p, q} \left(\mathbf{u}, \boldsymbol{\mu}, \boldsymbol{\sigma}, \theta \right)  \ | \ p + q \leq K \rbrace$ for a chosen $K \in \mathbb{N}$.

\bigskip
\noindent \textbf{Fully Trainable Gaussian Derivative Convolutional Layers.}
For all the configurations, the input and output tensors of the proposed layer as well as an intermediate tensor will respectively be denoted as $I$, $O$ and $J$ with $I \in \mathbb{R}^{l_{in} \times h_{in} \times f_{in}}$, $J, O \in \mathbb{R}^{l_{out} \times h_{out} \times f_{out}}$ and $l_{in}$, $h_{in}$, $l_{out}$, $h_{out}$ representing spatial dimensions and $f_{in}$, $f_{out}$ representing the numbers of feature maps of the tensors. The relations between $I$ and $J$ and between $J$ and $O$ as well as the relations between their dimensions will be given in this subsection.
Even though the contribution presented here can be seen as a set of Gaussian based layers, the primal aim is to propose a high level configurable layer which can be instantiated as any of the following configurations. The configurable levers are the following ones: i) the numbers of basis of Gaussian derivative kernels involved in the layer, ii) the number of linear combinations for each basis, iii) the type of linear combination and, iv) the orders of derivation of the Gaussian derivative kernels composing the bases.
\footnote{\url{https://github.com/Penaud-Polge/Fully-Trainable-Gaussian-Derivative-Layer}} 
\begin{table*}[t]
\centering
\small
\begin{tabular}{|c|c|c|c|c|c|c|}
 			\hline
 			& \multicolumn{6}{c|}{Layer Type} \\ \hline
			\# training samples& Free-Form & RFNN \cite{jacobsen2016structured} & DAU-ConvNet \cite{tabernik2016towards} & Ours & Ours-few & Ours-separated \\ \cline{1-7}
			1000 & 78.8 & 79.4 & 79.9 & \textbf{80.5} & 79.7 & 78.9  \\ \cline{1-7}
			5000 & 84.3 & 86.2 & 86.4 & 88.3 & 88.4 & \textbf{88.7}  \\ \cline{1-7}
			10000 & 86.4 & 88.4 & 89.1 & \textbf{90.5} & 90.1 & 89.3 \\ \cline{1-7}
			30000 & 87.7 & 90.4 & 90.9 &  92.2 & \textbf{92.4} & 91.4  \\ \cline{1-7}
			60000 & 89.3 & 91.2 & 92.1 &  \textbf{93.1} & 92.8 & 91.4 \\ \hline
\end{tabular}
\caption{Test accuracy obtained on the Fashion-MNIST dataset with different sizes of training dataset by the RFNN model, the DAU-ConvNet model, a usual free-form model and several configurations of the fully trainable Gaussian derivative network. All the networks have equivalent architectures and training conditions. The best testing accuracy is highlighted in bold.}
\label{tab1}
\end{table*}
We will go through all the configurations, beginning from the closest one to RFNN layers. We begin with a layer having only one fully trainable basis and using a standard linear combination. Its only basis will be referred to as $\lbrace \phi^{n} \left(\boldsymbol{\mu}^{n}, \boldsymbol{\sigma}^{n}, \theta^{n}\right) \ | \ n \in  [\![ 1,N ]\!] \rbrace$, where $N$ refers to the number of Gaussian derivative kernels composing the basis and where each $\phi^{n}$ is defined by \eqref{eq4}. Together with the basis, the layer also owns a set of coefficients $\lbrace \omega_{j,n}^{i} \ | \ n \in [\![ 1,N ]\!], i \in [\![ 1,f_{in} ]\!], j \in [\![ 1,f_{out}]\!] \rbrace$  used to compute the linear combination where the particular weight $\omega_{j,n}^{i}$ is used to compute the $j^{th}$ feature map of $J$ and is weighting the $n^{th}$ kernel of the basis to convolve the $i^{th}$ feature map of the input tensor $I$. In this case, the $j^{th}$ feature map of $J$ is computed by the following equation:
\begin{equation}
\label{eq6}
J_{j} = \sum_{i = 1}^{f_{in}}\left(\sum_{n = 1}^{N}\omega_{j,n}^{i} \phi^{n} \left(\boldsymbol{\mu}^{n}, \boldsymbol{\sigma}^{n}, \theta^{n}\right) \right) \ast I_i .
\end{equation}
Fig.\ref{fig1} (a) illustrates eq.\eqref{eq6} by schematically showing the relation between $I$ and $J$.
If, now, the layer has several bases, for example $B \in \mathbb{Z}^*_+$ many bases, we propose that each basis should be used to compute $\frac{f_{out}}{B} \in \mathbb{Z}^*_+$ feature maps of $J$. In order to differentiate each basis, the $b^{th}$ basis will be denoted as $\lbrace \phi^{b,n} \left(\boldsymbol{\mu}^{b,n}, \boldsymbol{\sigma}^{b,n}, \theta^{b,n}\right) \ | \ n \in  [\![ 1,N ]\!] \rbrace$ and its corresponding weights are denoted by $\lbrace \omega_{j,n}^{b,i} \ | \ n \in [\![ 1,N ]\!], i \in [\![ 1,f_{in} ]\!], j \in [\![ 1,\frac{f_{out}}{B}]\!] \rbrace$. Then, model from \eqref{eq6} becomes
\begin{equation}
\label{eq7}
J_{\left(b-1\right)\frac{f_{out}}{B} + j} = \sum_{i = 1}^{f_{in}}\left(\sum_{n = 1}^{N}\omega_{j,n}^{b,i} \phi^{b,n} \left(\boldsymbol{\mu}^{b,n}, \boldsymbol{\sigma}^{b,n}, \theta^{b,n}\right) \right) \ast I_i .
\end{equation}
In a policy of reducing the number of trainable parameters, in the same spirit as separable convolution \cite{haase2020rethinking}, we propose to separate the linear combination process. To describe this configuration, we will return to a layer having only one basis $\lbrace \phi^{n} \left(\boldsymbol{\mu}^{n}, \boldsymbol{\sigma}^{n}, \theta^{n}\right) \ | \ n \in  [\![ 1,N ]\!] \rbrace$. We propose to split the set of coefficients $\lbrace \omega_{j,n}^{i} \ | \ n \in [\![ 1,N ]\!], i \in [\![ 1,f_{in} ]\!], j \in [\![ 1,f_{out}]\!] \rbrace$ into two sets $\lbrace\lambda_{j,n} \ | \ n \in [\![ 1,N ]\!], j \in [\![ 1,f_{out} ]\!] \rbrace$ and $\lbrace\beta^{i}_{n} \ | \ n \in [\![ 1,N ]\!], i \in [\![ 1,f_{in}]\!] \rbrace$ and eq. \eqref{eq6} becomes
\begin{equation}
\label{eq9}
J_{j} = \sum_{n = 1}^{N} \lambda_{j,n}\left( \sum_{i = 1}^{f_{in}} \beta_{n}^{i} \phi^{n} \left(\boldsymbol{\mu}^{n}, \boldsymbol{\sigma}^{n}, \theta^{n}\right) \ast I_i\right)
\end{equation}
Fig.\ref{fig1} (b) illustrates this new configuration where an intermediate tensor is used before computing $J$. The split of the weights allows to decrease the number of weights from $f_{in} \times N \times f_{out}$ to $\left(f_{in} + f_{out}\right) \times N$. This configuration also offers the possibility to consider several bases.
In order to obtain $O$ from $J$, we first apply a batch normalization \cite{ioffe2015batch} to $J$. Then, one can apply a nonlinear function $F$ such as the rectified linear activation function or the sigmoid function which leads to $O = F\left(\gamma\frac{\left(J - \mu_{batch}\right)}{\sqrt{\sigma_{batch}^2 + \epsilon}} + \beta\right)$,
where $\gamma$ and $\beta$ are trainable parameters and where $\mu_{batch}$ and $\sigma_{batch}^2$ are either the mean and variance of the batch during training or the moving mean and the moving variance during inference. 
Finally, depending on the strides, a down-sampling operation can be applied, which gives: $l_{out} \leq l_{in}$, $h_{out} \leq h_{in}$.
\section{Experiments}
\label{sec:expe}
In this section we will first compare the proposed fully trainable Gaussian derivative layer to the ones involved in RFNN \cite{jacobsen2016structured} and in DAU-ConvNet \cite{tabernik2018spatially}. Then, we will use the proposed layer in two common architectures to show its relevance: as a first layer of a VGG16 model \cite{simonyan2014very} for cats and dogs classification, compared to a usual free-form convolutional layer and we will also use the proposed layer for a segmentation task with a fully trainable Gaussian derivative U-net and compare it with an equivalent classical U-net \cite{ronneberger2015u}.

\bigskip
\noindent \textbf{Comparison with related work.}
The aim of this experiment is twofold: First to compare several configurations of the proposed layer. More precisely, the different configurations will highlight the effect of using more or less trainable basis and the effect of the separation of the linear combination. Secondly, it will give a comparison between these different configurations and related work layers : the RFNN layer  \cite{jacobsen2016structured}, the DAU-ConvNet layer \cite{tabernik2016towards} and the usual free-form layer with filters of size $3\times3$. In order to demonstrate the positive effect of fully training the Gaussian derives, the values of the following parameters and quantities $\gamma$, $\mu_{batch}$ and $\sqrt{\sigma^2_{batch} + \epsilon}$, used in the batch normalization stage, are respectively fixed to one, zero and one for this experiment. Only $\beta$ is kept trainable, playing simply the role of a bias. For each type of layer, the same four-convolutional-layers architecture is used, ended by a global average pooling layer and a dense layer. The networks were trained using Fashion-MNIST \cite{xiao2017fashion} with different sizes of training dataset in order to study their behavior when a small amount of training samples is available. Each network was trained with a batch size of 32 and a cosine learning rate decay  defined as $lr_k = lr_0 \left(\alpha + \left(1 - \alpha\right)(1 + \cos\left(\frac{2 \pi \min\left(k, K_{max}\right)}{K_{max}} \right)\right)$. For RFNN, $\sigma = 1.5$ has been used for the first layer and $\sigma = 1$ for the others as they were the values used in \cite{jacobsen2016structured} for the MNIST dataset using a similar architecture. Concerning the DAU-ConvNet, $\sigma$ was let being learned by the network for each unit and the number of units for each filter took the value 10 as it corresponds to the number of filters composing the bases of the fully trainable Gaussian derivative network and RFNN. The networks representing three different configurations of the proposed layer are the following one: two networks, referred to as `Ours' and `Ours-few', using layers with non-separated linear combinations, one having half the number of basis (`Ours-few') than the other (`Ours') for each layer. And one network whose layers use separated linear combinations, with the same number of basis as `Ours', is referred to as `Ours-separated'. Three trials have been realized and the corresponding accuracy values have been averaged for the two smallest sizes of dataset in order to counter the effect of the random choice of images among the full training dataset. Results are presented in Tab.\ref{tab1} and show that the proposed layers obtain higher accuracy values compared to the other types of layers. The performance of `Ours-separated' highlights the benefits of the separated linear combination, as it allows comparable performances compared to the other configurations while decreasing the number of parameters. `Ours-separated' uses a third less parameters than `Ours'.

\bigskip
\noindent \textbf{VGG16.}
\begin{table}[t]
	\centering
	\small
	\begin{tabular}[aligm = m]{|c|c|c|}
		\hline
		Layer Type & Free-Form & Gaussian (Ours) \\ \hline
		Validation Accuracy & 97 & \textbf{98.1} \\ \hline
		Test Accuracy & 96.4 & \textbf{97.2}  \\ \hline
	\end{tabular}
	\caption{Accuracy obtained replacing and training the first layer of a pre-trained VGG16 on the Cats and Dogs dataset.}
	\label{tab2}
\end{table}
The purpose of this experiment is to study the behavior of the proposed layer when it is used as a first layer of a deep architecture. We considered a VGG16 pretrained on Imagenet \cite{deng2009imagenet}, and we perform transfer learning for Kaggle Cats and Dogs dataset. The obtained network achieves an accuracy of 92.8\% on the test dataset. We then replaced the first layer of the network by a fully trainable Gaussian derivative layer, froze the rest of the network and trained it. Results presented in Tab.\ref{tab2} show that the proposed layer is adapted to be used as the first layer of a convolutional neural network as it obtains better performances compared to a classical free-form layer. The next experiment will investigate the legitimacy of using an architecture using exclusively the proposed layer.

\begin{figure}[b!]
	\centering
	\bgroup
    \def\arraystretch{0.4}
    \setlength\tabcolsep{0.4pt}
	\begin{tabular}{c c c c c}
	\includegraphics[width=0.19\linewidth]{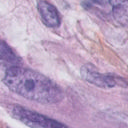}&
	\includegraphics[width=0.19\linewidth]{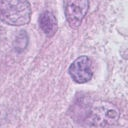}&
	\includegraphics[width=0.19\linewidth]{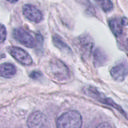}&
	\includegraphics[width=0.19\linewidth]{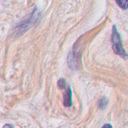}&
	\includegraphics[width=0.19\linewidth]{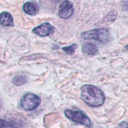} \hfill \\
	\includegraphics[width=0.19\linewidth]{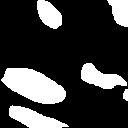}&
	\includegraphics[width=0.19\linewidth]{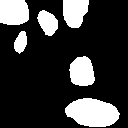}&
	\includegraphics[width=0.19\linewidth]{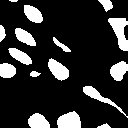}&
	\includegraphics[width=0.19\linewidth]{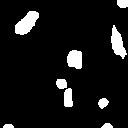}&
	\includegraphics[width=0.19\linewidth]{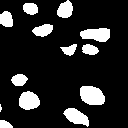} \hfill \\
	\includegraphics[width=0.19\linewidth]{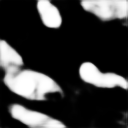}&
	\includegraphics[width=0.19\linewidth]{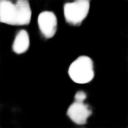}&
	\includegraphics[width=0.19\linewidth]{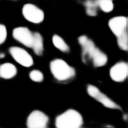}&
	\includegraphics[width=0.19\linewidth]{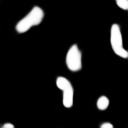}&
	\includegraphics[width=0.19\linewidth]{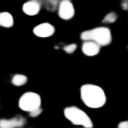} \hfill \\
	\includegraphics[width=0.19\linewidth]{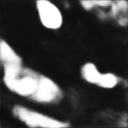}&
	\includegraphics[width=0.19\linewidth]{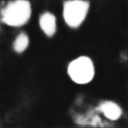}&
	\includegraphics[width=0.19\linewidth]{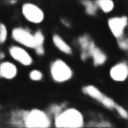}&
	\includegraphics[width=0.19\linewidth]{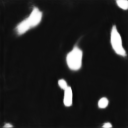}&
	\includegraphics[width=0.19\linewidth]{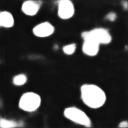} \hfill
	\end{tabular}
	\egroup
\caption{First row: Input images. Second row: Ground truths. Third row: Classical U-net predictions. Fourth row: Gaussian U-net predictions.}
\label{fig4}
\end{figure}

\bigskip
\noindent \textbf{Gaussian U-net.}
\begin{table}[t]
	\centering
	\small
	\begin{tabular}[aligm = m]{|c|c|c|}
		\hline
		Network Type & U-net & Gaussian U-net \\ \hline
		Test Accuracy & \textbf{94.7} & 94.4 \\ \hline
		\# Trainable parameters & 97\,202 & \textbf{25\,298} \\ \hline
	\end{tabular}
	\caption{Accuracy obtained on the histopathology test dataset and \# of trainable parameters of Gaussian and classical U-net.}
	\label{tab3}
\end{table}
This experiment aims to test the fully trainable Gaussian derivative layer for a segmentation of nuclei in histopathology images. The used dataset has been generously provided by the authors of \cite{naylor2017nuclei}. It is composed of 50 images of size~$512\times512$~with their ground truths. For each of the 50 images, 16 images of size $128\times128$ have been extracted, forming a dataset of 800 images with their associated ground truths (600 for training, 200 for testing). We compare the performance of a classical U-net (with filters of $5\times5$) by replacing the convolutional layers by our proposed Gaussian layer, called Gaussian U-net.
Results in Tab.\ref{tab3} shown that the fully trainable Gaussian derivative layer can compete with classical free-form layers while using nearly 25\% of parameters. Fig.\ref{fig4} shows an example of predictions for both U-nets, highlighting they are quite similar despite the reduction in the number of parameters.


\section{Conclusion}
\label{sec:conclu}
We presented a configurable fully trainable Gaussian derivative layer generalizing previous works while keeping their advantages. It allows for more diversified shapes but still with the possibility to reduce the number of parameters. The conducted experiments showed the behavior of the proposed layer, its competitiveness with related previous work and that it offers enough diversity to deal with challenging tasks in computer vision.
\newpage
\bibliographystyle{IEEEbib}
\bibliography{ICIP}
\end{document}